\documentclass{article}

\usepackage{microtype}
\usepackage{graphicx}
\usepackage{subfigure}
\usepackage{booktabs} 

\usepackage{hyperref}

\usepackage[accepted]{icml2023}

\usepackage{amsmath}
\usepackage{amssymb}
\usepackage{mathtools}
\usepackage{amsthm}
\usepackage{multicol}
\usepackage[capitalize,noabbrev]{cleveref}

\theoremstyle{plain}

\theoremstyle{definition}

\theoremstyle{remark}

\usepackage[textsize=tiny]{todonotes}

\icmltitlerunning{}

\begin{document}

\twocolumn[
\icmltitle{Gradient-Free Textual Inversion}




\begin{icmlauthorlist}
\icmlauthor{Zhengcong Fei}{comp}
\icmlauthor{Mingyuan Fan}{comp}
\icmlauthor{Junshi Huang}{comp,*}
\end{icmlauthorlist}

\icmlaffiliation{comp}{Meituan, Beijing}
\icmlaffiliation{*}{Corresponding author}

\icmlcorrespondingauthor{Zhengcong Fei}{feizhengcong@meituan.com}
\icmlcorrespondingauthor{Junshi Huang}{huangjunshi@meituan.com}

\icmlkeywords{Machine Learning, ICML}

\vskip 0.3in
]



\printAffiliationsAndNotice{}  

\begin{abstract} 
Recent works on personalized text-to-image generation usually learn to bind a special token with specific subjects or styles of a few given images by tuning its embedding through gradient descent. It is natural to question whether we can optimize the textual inversions by only accessing the process of model inference. As only requiring the forward computation to determine the textual inversion retains the benefits of less GPU memory, simple deployment, and secure access for scalable models. In this paper, we introduce a \emph{gradient-free} framework to optimize the continuous textual inversion in an iterative evolutionary strategy. Specifically, we first initialize an appropriate token embedding for textual inversion with the consideration of visual and text vocabulary information. Then, we decompose the optimization of evolutionary strategy into dimension reduction of searching space and non-convex gradient-free optimization in subspace, which significantly accelerates the optimization process with negligible performance loss. Experiments in several applications demonstrate that the performance of text-to-image model equipped with our proposed gradient-free method is comparable to that of gradient-based counterparts with variant GPU/CPU platforms, flexible employment, as well as computational efficiency.


\end{abstract}

\section{Introduction}

Large-scale text-to-image models, enabling high-quality and diverse synthesis of images based on a text prompt written in natural language, have achieved remarkable progress and become an exciting direction \cite{nichol2021glide,saharia2022photorealistic,ramesh2022hierarchical,rombach2022high,yu2022scaling}. 
One of the main advantages of these models is the strong semantic prior learned from scalable collections of image-caption pairs, leading to their broad application in artistic creation, \emph{e.g.}, as sources of inspiration, and even in the designing of new physical products.
While the generation capabilities of text-to-image models are unprecedented, they lack the ability to mimic the appearance of subjects in a given reference set, and synthesize novel renditions of the same subjects in different contexts \cite{ruiz2022dreambooth}, \emph{i.e.}, even the most detailed textual description of an object may yield instances with different appearances \cite{gal2022image}.

Personalization of text-to-image generation is proposed to address this kind of issue to certain extent. The general idea is to expand the embedding dictionary of text encoder by adding a new concept token of specific subject or style which the users want to generate. 
In particular, textual inversion \cite{gal2022image,daras2022multiresolution} is a powerful technique that can learn the new pseudo token in the embedding space for the representation of new concept. Remarkably, this tuned token can be composed in language to produce kinds of creative compositions.
Though textual inversion keeps the major text-to-image model unchanged, optimizing the parameters of pseudo token still requires back-propagation through the entire model, which is expensive or even unfeasible for many applications with limited resource.
Recently, it has been demonstrated that scaling up the model size is promising to achieve better semantic understanding \cite{yu2022scaling,yu2022coca}, while the growing model size leads to an increment in tuning cost as well as unstable fine-tuning process. 

To make personalized text-to-image paradigm benefiting a wider range of audiences, a natural question raises: 
\emph{Can we optimize the specific textual inversion when we only have access to the inference of text-to-image model?} 
In such scenario, users cannot access the derivatives or adjust the parameters of text-to-image model but accomplish the text inversion to obtain an object or style of interest bounded by a range of inferences. 
In contrast to gradient-based optimization, the gradient-free framework can be highly optimized by acceleration tools such as ONNX and TensorRT.
In addition, the optimization of textual inversion can be decoupled from the complicated deployment of scalable training framework.
Although solving optimization problems in an inference-only setting is considerably challenging \cite{wang2018stochastic}, our gradient-free framework introduces a new and effective paradigm of personalized text-to-image generation.

Here, we resort to the gradient-free optimization (GFO), also termed as black-box, zeroth-order or derivative-free optimization \cite{conn2009introduction,kolda2003optimization,rios2013derivative,sahu2019towards}.
In general, GFO involves a kind of optimization algorithms that do not require gradients, but only rely on function values or fitness values of iteratively sampled solutions \cite{rios2013derivative}. 
However, GFO algorithms are known to suffer from a slow convergence in high-dimensional search space, 
due to the massive searching directions for continuous text embedding.
To alleviate the searching problem in textual inversion, we propose a composing initialization strategy to effectively reduce the exploration cost. 
Moreover, inspired by the recent works that common pre-training models, despite their large number of parameters, have a very low intrinsic dimensionality \cite{aghajanyan2021intrinsic,qin2021exploring}. That means, there exists a low-dimensional subspace that is as effective for tuning as the full dimension space. 
Therefore, with appropriate subspace decomposition in objective function, the textual inversion optimization can be effectively solved in low-dimensional subspace.

Based on these insights, this paper presents a \textbf{gradient-free} framework to solve the personalized text-to-image generation task. 
Specifically, we manage to optimize the pseudo-token embedding given several images by iteratively forwarding the text-to-image model and design the loss function to measure fitness of sampled solutions.
To improve the convergence and stability of optimization, we introduce to (\textbf{i}) initialize the pseudo-token embedding with general condition, \emph{i.e.}, non-parametric cross-attention of pre-trained word embedding and personalized visual features; 
(\textbf{ii}) decompose the original searching space of GFO into a smaller subspace using Principal Components Analysis (PCA) or prior normalization and solve the transferred problem with some derivative-free optimizer in the subspace for incremental elements. 
In particular, we adopt Covariance Matrix Adaptation Evolution Strategy (CMA-ES) to search the target embedding by exploration and exploitation in a parameterized search distribution.
Encouragingly, this gradient-free textual inversion allows users to optimize their specific demand locally on the resource-limited devices even without GPUs. 
We use the stable diffusion model \cite{rombach2022high} as the base model in our experiments, though our method is not constrained to any specific text-to-image models. 
Experiment results on several tasks demonstrate that gradient-free optimization achieves compariable performance with its gradient-based counterparts in terms of quantitative analyses and human evaluation. 

Overall, the contributions of this paper are four fold: 
\begin{itemize}
    \item We introduce a new scenario of textual inversion in gradient-free framework, which to our best knowledge is the first trial of GFO methods on personalized text-to-image generation tasks;
    \item This paper offers a solution with an improved evolution strategy in the searching scenario to accomplish the common text-to-image personalization task;
    \item To accelerate the convergence of iterative process, we provide the general condition initialization for pseudo-token embedding and decomposed subspace for effective incremental optimization;
    \item Empirical results show that gradient-free textual inversion can successfully deal with real-world applications, achieving comparable performance with gradient-based counterparts. 
    The source code will be publicly available. 
\end{itemize}

\section{Related Works}

\paragraph{Text-to-image synthesis.}
Text-to-image synthesis, \emph{i.e.}, generation of images conditioned on text prompt, have been extensively studied in recent years \cite{qiao2019mirrorgan,li2019controllable,ding2021cogview,hinz2020semantic,tao2020df,li2019object,qiao2019learn,ramesh2021zero,zhang2018photographic,crowson2022vqgan,gafni2022make,chang2023muse}. In particular, large-scale diffusion-based models \cite{saharia2022photorealistic,ramesh2022hierarchical,yu2022scaling,ding2022cogview2} demonstrated an astonishing semantic conditioned generation. 
However, these models do not provide a fine-grained control cue over a generated image. That is, given text descriptions, it is hard to preserve the identity of a subject consistently across different situations, since modifying the context in the prompt also modifies the appearance of the subject. 
In this regard, \cite{liu2021more} propose a modification to classifier guidance that allows for the guidance of diffusion models using images and text, allowing for semantic variations of an image, although the identity of the subject often varies. To overcome subject modification, several works \cite{nichol2021glide,avrahami2022blended,couairon2022diffedit} assume that the user provides a mask to restrict the area in which the changes are applied. Recent work of prompt-to-prompt \cite{hertz2022prompt} allows for local and global editing without any input mask requirements, but meets hard up when given image set and personalizes to new situations.

\paragraph{Inversion of diffusion models.}
Diffusion models inversion, which aims to find a noise map and a conditioning vector that corresponds to a generated image, is a challenging task and provides a promising solution to image generation controls \cite{wu2022unifying,mokady2022null}. 
It is well known that addition of slightly different noise to an image followed by denoising may yield a completely different image. 
To this end, \cite{choi2021ilvr} tackle inversion by conditioning the denoising process on noised low-pass filter data from the target image. \cite{dhariwal2021diffusion} show that deterministic DDIM sampling \cite{song2020denoising} can be inverted to extract a latent noise map that will produce a given real image. 
\cite{ramesh2022hierarchical} use this method to generate cross-image interpolations or semantic image editing using CLIP latent vectors. However, all these methods face limitations in generating novel conditions of a subject while preserving fidelity. 
This paper is based on textual inversion \cite{gal2022image}, which proposes a  method that learns to represent visual concepts, like an object or a style, through new pseudo-words in the embedding space of a frozen text-to-image model. Their approach searches for the optimal embedding that can represent the concept with gradient backward and calibrated learning rate scheme \cite{ruder2016overview}. In contrast, we try to optimize the inversion embedding without gradient, enabling the generation of novel images of the subject while preserving computation efficiency and safety.

\paragraph{Personalization.}
In recent years, personalization has become a prominent factor in various fields within machine learning area \cite{yoganarasimhan2020search,schneider2019personalized}.
Within the vision and graphics community, there are few works that tackle the problem of novel synthesis of subjects using GANs. For example, \cite{casanova2021instance} propose a method to condition GANs on instances, such that variations of the same instance can be generated. 
However, the generated subjects share features with the conditioning instance, and are not identical, thus cannot solve the personalization image generation problem. 
\cite{nitzan2022mystyle,melnik2022face} finetune a face synthesis GAN on a specific identity to build a personalized prior. 
While it requires hundreds of images to learn a semantic prior only for the facial domain, this paper aims to reconstruct the identity of different types or styles of subjects in new contexts with only several casual images and can be transferred with prompts \cite{gal2022image,ruiz2022dreambooth}.

\paragraph{Gradient-free optimization.}
Gradient-free optimization (GFO) realizes optimization only via the function values on the sampled solutions. Most GFO algorithms share a common structure of sampling and updating to enhance the quality of solutions.
Representative GFO algorithms include evolutionary algorithms \cite{hansen2001completely}, Bayesian optimization \cite{frazier2018tutorial}, etc.
They initially sample some random solutions in the search space and evaluate them. The model is able to imply the regions which may contain some potentially better solutions with evaluated training objectives. Finally, they then sample the new solutions from that model and update the model. GFO algorithms repeat the above sampling-and-updating procedure so as to iteratively enhance the quality of solutions \cite{rechenberg1989evolution}. 
Due to their ability to address complex optimization tasks, GFO algorithms have achieved many impressive applications in automatic machine learning \cite{snoek2012practical}, reinforcement learning \cite{salimans2017evolution,hu2017sequential}, objective detection \cite{zhang2015improving}, few-shot learning \cite{sun2022black,xu2022gps,chai2022clip}, and black-box attack \cite{alzantot2019genattack}.

\begin{figure*}[t]
\centering
\includegraphics[width=1.9\columnwidth]{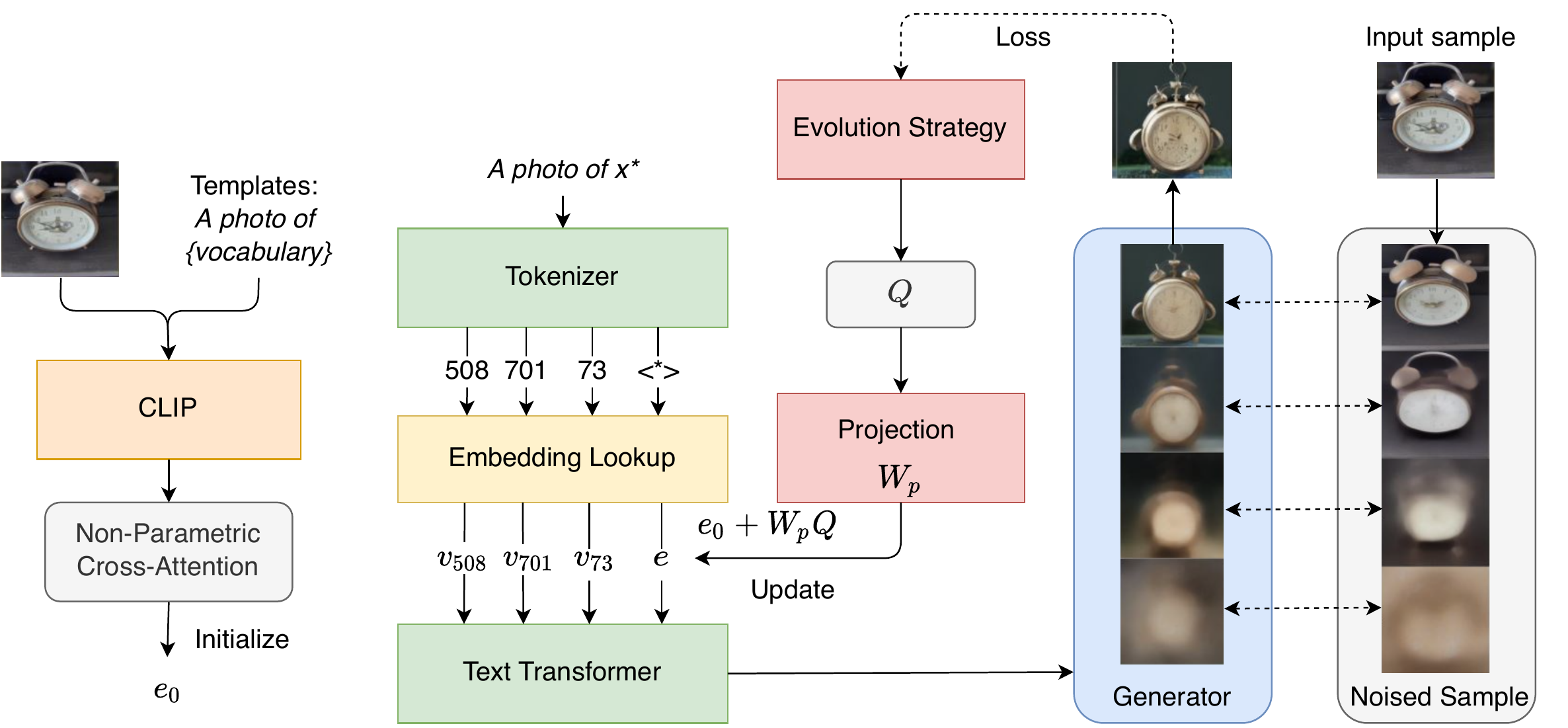} 
\caption{\textbf{Overview of gradient-free textual inversion framework for personalized text-to-image generation.} Specifically, the evolution strategy is performed iteratively to explore and exploit pseudo-token embedding. To accelerate the optimization, ($\textbf{i}$) the textual inversion is initialized with weighted cross-attention between given images and vocabulary, and ($\textbf{ii}$) optimization is conducted in a decomposition subspace through PCA or prior normalization.
}
\label{fig:framework}
\end{figure*}

\section{Approach}

\subsection{Formulation}


This work is based on a diffusion-based text-to-image model, which is trained to restore a clean image $Y$ from a corrupted observation $z_t(Y)$ with text condition $X$.
The corruption, which typically is a Gaussian noise, can be applied either on the raw image \cite{ramesh2022hierarchical,saharia2022photorealistic}, or the latent space of image encoding \cite{rombach2022high}. 
We denote $z_t(Y)$ as the diffused observation of image $Y$ at time $t$ with a preset schedule. 
When denoising from $z_t(Y)$, the tokenized text $X$ is encoded into latent variables by the text encoder, and then integrated into the denoising network for noise prediction. 

Textual inversion \cite{gal2022image} considers the task of finding an embedding $e^*$ for a pseudo token $x^*$ that represents the concept described by a small collection of images $\mathcal{Y}=\{Y_1, \ldots, Y_N\}$. The optimal embedding can be obtained with reconstruction loss:
\begin{equation}
\begin{split}
    L(e) = \mathbb{E}_{t \in \mathcal{U}(0,1)} & \mathbb{E}_{Y \sim \mathcal{Y}} \mathbb{E}_{z_t \sim q_t(z_t|Y,t)} \\ 
    & || \hat{\epsilon}(z_t(Y), t, f(X)) - \epsilon(z_0, z_t) ||^2, \label{eq:denoiseloss}
\end{split}
\end{equation}
where $X$ is a specific text prompt containing $x^*$, $q_t$ is the distribution of diffused observations, $\hat{\epsilon}$ is the denoising network, $f$ is the text encoder, $\epsilon$ is the unscaled noise sample, 
and $e$ is the optimizing pseudo-token embedding of $x^*$. Intuitively, keeping the pre-trained denoising network $\hat{\epsilon}$ and text encoder $f$ fixed, textual inversion learns to capture the unique visual concept $x^*$ in input images by optimizing its embedding $e$.


\subsection{Gradient-Free Textural Inversion}
In this study, we aim to determine the optimal pseudo-token embedding $e^\star$ in a gradient-free scenario, \emph{i.e.}, the parameters of text-to-image models are not available to the optimizer but can only be accessed in inference. 
However, directly using a gradient-free optimization strategy is intractable, as the dimensionality of the embedding $e \in \mathbb{R}^D$ can be large.
To handle this high-dimensional optimization, we manage to decompose the embedding into $e = e_0 + W_p Q$ based on a well-initialized embedding $e_0$ and search the increment of $Q$ in a low-dimensional subspace. 
Thus, our objective function becomes: 
\begin{align}
    Q^\star = \mathop{\arg\min}_{Q \in \mathcal{Q}}L(e_0 + W_p Q)\,, \label{eq:loss}
\end{align}
where $\mathcal{Q}$ is the sub-search-space. As illustrated in Figure \ref{fig:framework}, our framework includes two stages:  (\textbf{i}) initialize the pseudo-token embedding $e_0$ with a reliable strategy; (\textbf{ii}) optimize textual inversion based on the incremental $Q \in\mathbb{R}^d$ with an iterative evolution strategy in a smaller subspace reduced by linear projection matrix $W_p \in\mathbb{R}^{D\times d}$, where $d\ll D$.
Note that to ensure the loss value $L$ accurately reflect the optimization direction, it is necessary to fix the added noise level, \emph{i.e.}, randomly selected parameter $t$, in each iteration during the evaluation.

In the following, we will specify the advanced evolution strategy to optimize incremental part $Q$ progressively in Sec. \ref{sec:evolution}, and obtain the initialization of inversion embedding $e_0$ in Sec. \ref{sec:initialization}. Finally, we propose two subspace decomposition strategies for projection $W_p$, including PCA and prior normalization, in Sec. \ref{sec:subspace}.


\begin{figure*}[h]
\centering
\includegraphics[width=2\columnwidth]{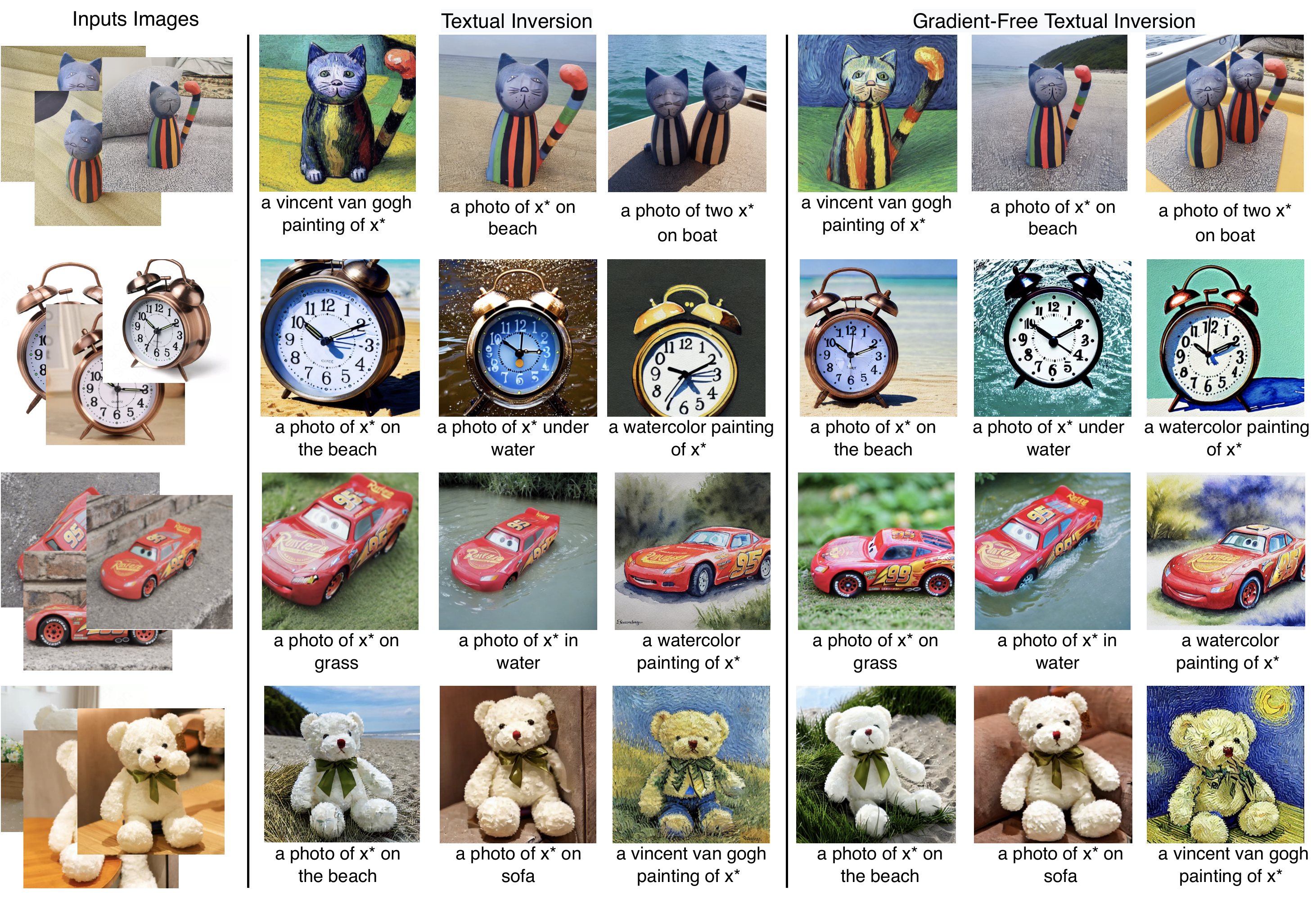} 
\caption{\textbf{Personalized text-guided image generation.} It demonstrates that with gradient-free optimization method, we can use the pseudo-token for the learned concept to create personalized images as if it was a normal word token. Importantly, our method performs on par and sometimes better than standard textual Inversion on this task.}
\label{fig:text_guid}
\end{figure*}

\begin{figure*}[t]
\centering
\includegraphics[width=2\columnwidth]{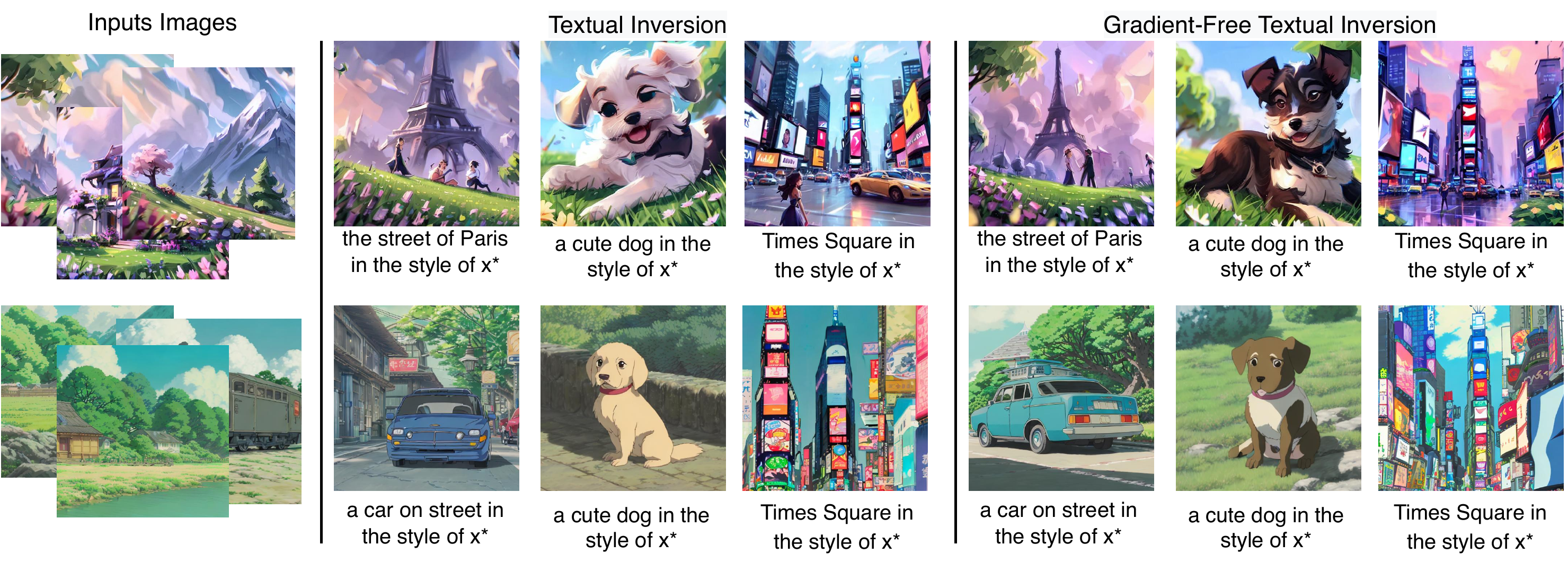} 
\caption{\textbf{Personalized style-guided image generation.} As the textual-embedding space can represent more abstract concepts, including different art styles, we can also discover that pseudo-token embedding with gradient-free optimization can represent the style of given images powerfully.}
\label{fig:style}
\end{figure*}

\subsection{Evolution Strategy} 
\label{sec:evolution}

To better optimize gradient-free paradigm, we adopt the Covariance Matrix Adaptation Evolution Strategy (CMA-ES) \cite{hansen2001completely,hansen2003reducing}, which is a widely used evolutionary algorithm for non-convex gradient-free optimization in continuous domain.
Generally, during CMA-ES, a parameterized search distribution model, \emph{i.e.}, multivariate normal distribution, is maintained for the exploration of candidate solutions with adaptive covariance matrix. 
With objective function for fitness evaluation, the CMA-ES exploits the optimal solutions with top fitness values in each iteration.
By transferring the CMA-ES into subspace domain, the candidate update of subspace pseudo-token embedding at $t$-th step can be explored as: 
\begin{equation}
    {Q}_i^{t+1} \sim {m}^{t} + {\sigma}^{t}\mathcal{N}({0}, {C}^{t})\,,
\end{equation}
where $i=1,\dots,K$ and $K$ is the population size. ${m}^{t}\in \mathbb{R}^d$ is the mean vector of the search distribution at iteration step $t$, $\sigma ^{t}$ is the overall standard deviation that controls the step length, and ${C}^{t}\in \mathbb{R}^{d\times d}$ is the covariance matrix that determines the shape of the distribution ellipsoid. 
In exploitation, each ${Q}_i^{t+1}$ is projected into the original text embedding space and calculate the fitness value according to Equation \ref{eq:loss}. 
Finally, by maximizing the likelihood of successful trials,  ${m}^{t}$, $\sigma^{t}$, and ${C}^{t}$ are updated accordingly. 
In the continuous exploration and exploitation processes, we maintain a global optimal $Q^\star$ as result. 
Please refer to \citet{hansen2016cma} for more details.

\subsection{General Conditioned Initialization}
\label{sec:initialization}

In our preliminary experiments, we find that both the convergence and final performance of gradient-free optimization are greatly impacted by the quality of embedding initialization. 
Instead of using manual token initialization in \cite{gal2022image}, we introduce an adaptive initialization method based on the non-parametric cross-attention. 

Specifically, for each image $Y_i \in \mathcal{Y}$ and text prompts $X$, \emph{e.g.}, “a photo of $v$” for objective and “a style of $v$” for style with every token $v$ in vocabulary $\mathcal{V}$, we use the pre-trained CLIP model $c(\cdot)$ \cite{radford2021learning} to extract their features and calculate the cosine similarity $s(\cdot)$ between every pair of visual and text embedding. Then, the similarity scores are used as embedding weights to get the initialization embedding of pseudo-token as:
\begin{equation}
    e_0 = \frac{1}{N} \sum_{i=1}^N \sum_{v \in \mathcal{V}}  ~ \frac{e^{s(c(Y_i), c(X_{v}))}}{\sum_{v\prime \in \mathcal{V}}  e^{s(c(Y_i), c(X_{v\prime}))}} ~  e_v, 
\end{equation}
where $e_v$ is the text embedding of token $v$. 
Note that 
($\textbf{i}$) as the CLIP model and diffusion-based text-to-image model  \cite{rombach2022high} use the same pre-trained text encoder, the initialized embedding $e_0$ is constructed under the same distribution of latent space $\mathcal{E}$ 
and ($\textbf{ii}$) the resulting embedding can be viewed as an visual-adaptive integration of existing vocabulary and thus contains sufficient visual semantic and explainable representation.


\subsection{Subspace Decomposition}
\label{sec:subspace}

We finally shed light on the subspace projection $W_p$. 
Assume that the intrinsic dimensionality of the personalized text-to-image model is $d^\star$, we believe that there exists a $Q \in \mathbb{R}^d$ such that $L(e)  \approx L(e_0 + W_p Q)$ as long as $d\ge d^\star$.
Similarly, previous works in high-dimensional GFO \cite{wang2016bayesian,qian2016derivative,letham2020re} try to utilize a normal distribution to set each entry in subspace projections. However, they usually simply use $\mathcal{N}(0,1)$ or $\mathcal{N}(0, \frac{1}{d})$ to randomly initialize the projection weight, both of which under-perform in our textual inversion scenario. To this end, we propose two decomposition strategies:

\paragraph{Principal components analysis.}

PCA is predominantly used as a dimensionality reduction technique in domains like facial recognition, image classification and image compression \cite{shlens2014tutorial}.
Generally, we gather all the text embedding in vocabulary of text encoder as training data and calculate the PCA on this data.
Then, we use the first $d$ eigenvectors to build the projection matrix $W_p$.
In this way, the optimized incremental parts can be inverse-transformed into the original space by rule and line.

\paragraph{Prior normalization.} 
Inspired by \cite{xu2022gps,chai2022clip}, we take into account the distribution of word embeddings of textual encoder as prior information to set normalization-based projections. Specifically, we use the modified normal distribution with standard deviation as: 
\begin{equation}
    \sigma_{p} = \lambda \frac{\sigma_e }{\sqrt{d} \sigma_{Q}}, \label{eq:sigma}
\end{equation}
where $\sigma_e$ is the observed standard deviation of text embedding, $\sigma_Q$ is the standard deviation of the normal distribution maintained by CMA-ES, and $\lambda$ a is constant scalar to stretch the distribution. Initially, we set $\mu_Q=\mu_p=0$ so that no prior knowledge about the optimization direction is incorporated. 
The main idea behind the above calculation is to match the distribution, \emph{i.e.}, variance, between the projected pseudo-token embedding and word embeddings. A detailed derivation of Equation \ref{eq:sigma} is provided in Appendix~\ref{sec:append_A}.

\section{Experiments}

\subsection{Experimental Settings}
We retain the original hyper-parameter choices of the stable diffusion model from diffusers \cite{von-platen-etal-2022-diffusers} and the CLIP model from huggingface \cite{wolf2020transformers}. 
All of our experiments are conducted on 32G-V100 GPUs. 
For CMA-ES, we set the popsize to 30 and all results are produced using 8, 000 $\sim$ 13,000 exploration steps with a subspace dimensionality of 256. We find that these parameters work well for most personalization cases. We also note that better results can be achieved with different optimization steps similar to \cite{gal2022image}. 
For the given image set, we resize all images to $512 \times 512$ and repeat them for 10 times with different text prompts. The evaluation batch size for each text embedding is set to 20, which is much larger than that of 1 in standard textual inversion. 
For simplicity, we reference the personalied subject or style unique identifier as $x^\star$. 

\subsection{Qualitative Comparisons and Applications}

In this section, we demonstrate a range of personalization applications enabled through gradient-free textual inversion, and provide visual comparison to the standard textual inversion, from both text-guided image synthesis and style transfer perspectives.

\paragraph{Text-guided image synthesis.}

To evaluate the ability of composing new scenes or formations by incorporating the learned pseudo-tokens into new conditioning texts, for each concept, we show exemplars for input images, along with an array of generated images and their conditioning texts for both standard textual inversion as well as our gradient-free inversion. Figure \ref{fig:text_guid} depicts the cases.  From these exemplar results, it is easy to see that both textual inversions are able to jointly reason over both the new concepts and their large body of prior knowledge, bringing them together in a different creation. 
More encouragingly, despite the fact that our optimization is conducted in an embedding subspace, it still captures the semantic concepts and subjective details effectively. For example, observe the car's features (row three) to maintain the specific patterns and the different counting numbers of target toys (row one). Finally, compare with gradient-based inversion with a well-designed learning process, our proposed gradient-free method can adaptively balance the exploration and exploitation, and provide a safe and convenient usage.

\paragraph{Style transfer.}

\begin{figure}[t]
\centering
\includegraphics[width=1\columnwidth]{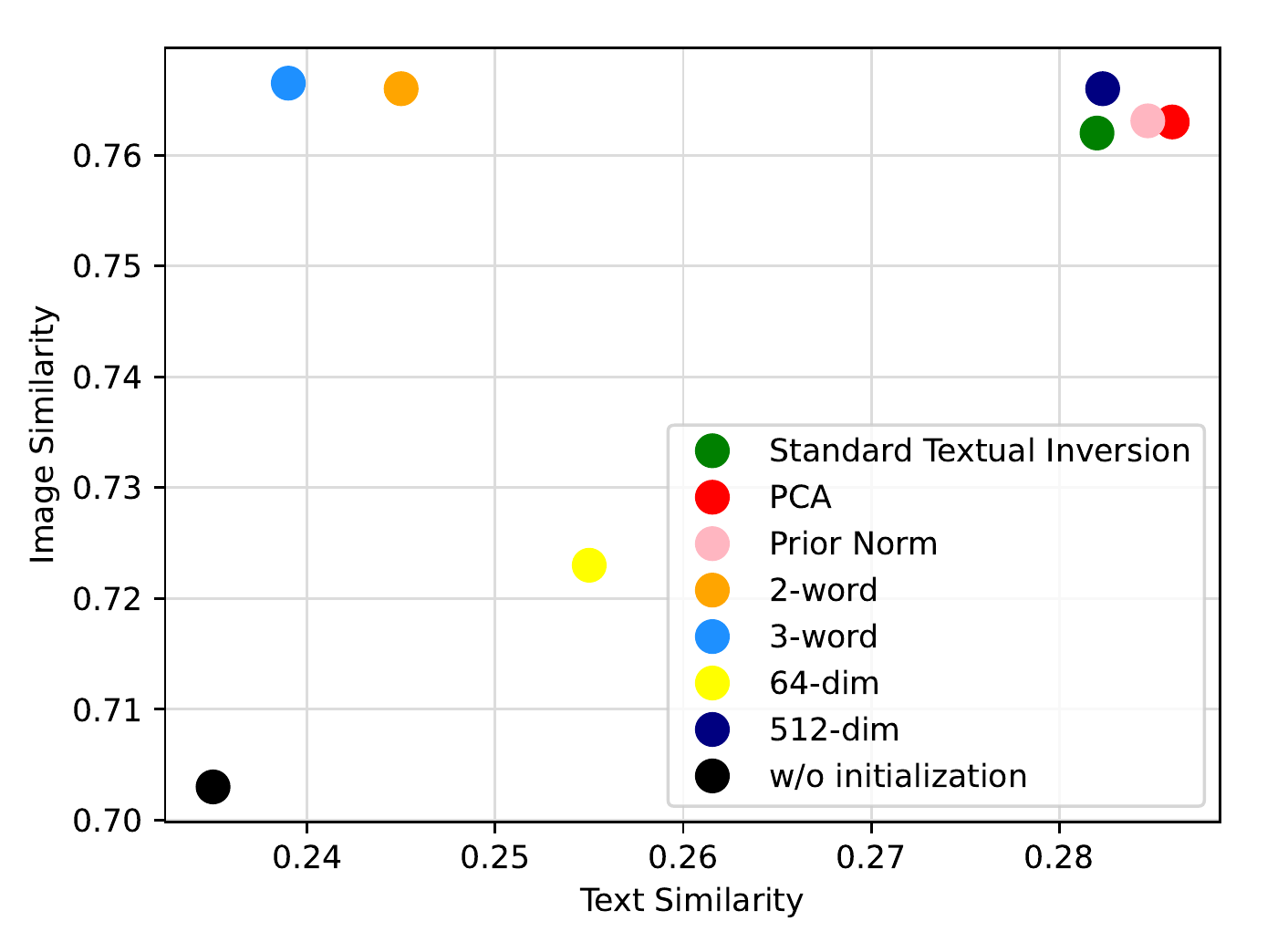} 
\caption{\textbf{Quantitative analysis in CLIP-based evaluations} compared with standard textual inversion and gradient-free inversion variants including different pseudo token numbers and decomposition subspace dimensionality.}
\label{fig:ablation}
\end{figure}

\begin{figure*}[t]
\centering
\includegraphics[width=2\columnwidth]{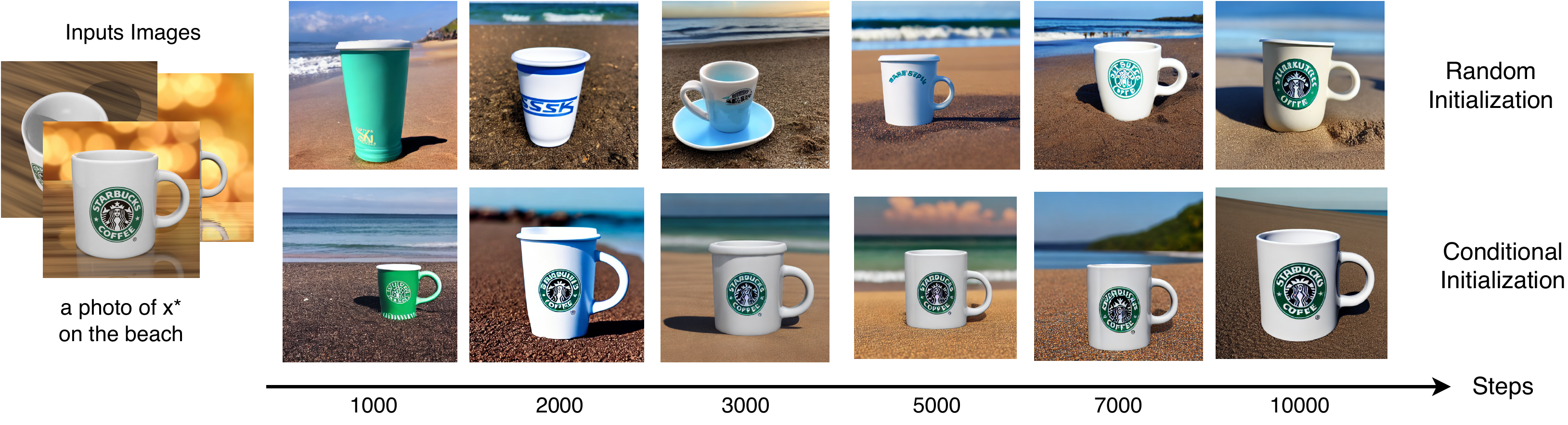} 
\caption{\textbf{Effect of general condition initialization.} We can see that cross-attention method provides a good initialization point for special pseudo-token embedding and obtains a prominently faster convergence.}
\label{fig:initialization}
\end{figure*}

Style-guided image generation, where users target to capture a unique style of a specific artist based on several painting works and apply it to new content, is an important application for personalized text-to-image generation. 
Note that different from traditional style transfer, we do not necessarily wish to maintain the content of some input images. Instead, we offer the text-to-image model sufficient freedom to decide how to plot the subject according to the input prompt and merely required an appropriate style. 
Specifically, as shown in Figure \ref{fig:style}, we give the model several images of various content with a shared style, and equipped with the training texts with prompts of the template form, \emph{e.g.}, “A painting in the style of $x^\star$”. Through an iterative evolution process, our gradient-free model can find a pseudo-token that represents a specific, implicit style. 
As expected, we can see that both textual inversion methods hold the ability to capture concepts that extends beyond simple object reconstructions and into more abstract ideas.

\subsection{Quantitative Analysis} 

\paragraph{Evaluation metrics.}
Following \cite{gal2022image,nichol2021glide}, we consider estimating the inversion performance in two fronts: ($\textbf{i}$) \emph{reconstruction}: to replicate the target concept, \emph{i.e.}, measure the similarity by considering semantic CLIP-space distances. 
Specifically, for each concept, we generate 64 images with the prompt: “A photo of $x^\star$”. The reconstruction score is then the average pair-wise CLIP-space cosine-similarity between the generated images and the images of the concept-specific training set; 
($\textbf{ii}$) \emph{editability}: to evaluate the ability to modify the concepts using textual prompts. 
Specifically, we produce a set of images using prompts of varying condition settings. These range from background modifications, \emph{e.g.}, “A photo of $x^\star$ on the moon”, to style changes, \emph{e.g.}, “An oil painting of $x^\star$”.
For each prompt, we generate 64 image samples, calculate the average CLIP-space embedding of the samples, and compute their cosine similarity with the CLIP-space embedding of the textual prompts, where we omit the placeholder $x^\star$, \emph{i.e.}, “A photo of on the moon” and “An oil painting”. 
Note that a higher editability score indicates better editing capability and more faithfulness to the prompt itself.

\paragraph{Baselines. }
We adopt the variants including: 
($\textbf{i}$) standard textual inversion \cite{gal2022image} that optimizes the inversion with gradient-based optimizer; 
as well as gradient-free inversion variants: 
($\textbf{ii}$) gradient-free methods that optimized with different subspace decomposition strategies, denoted as PCA and Prior Norm for simplicity.
($\textbf{iii}$) extended inversion token number that considers an extension to two and three pseudo-words, denoted as 2-word and 3-word, respectively. 
($\textbf{iv}$) various subspace dimensionality that employs different subspace dimensionality, denoted as 64-dim and 512-dim.
($\textbf{v}$) gradient-free method without conditional initialization with the same optimization steps, \emph{i.e.}, initialize with random token embedding, denoted as w/o init.

To better evaluate the performance, we additionally collect 50 concepts from open source data online and carefully selected by well-educated people to ensure the quality, which will be publicly available to improve reproducibility and foster new research in the field. The evaluation results of all gradient-free variants as well as standard textual inversion are summarized in Figure \ref{fig:ablation}.

\paragraph{Effect of inversion token number.} 
We illustrate the performance of different pseudo-token number to clarify its influence.
According to the evaluated results, we find that, on one hand, the overall performances of 2-3 pseudo-tokens are close; On the other hand, the single-word gradient-free method gets comparable reconstruction quality, while considerably improved editability over all multi-word baselines. We attribute that incorporating more tokens may distract and hurt the textual prompt guiding. 
Consistent with previous work \cite{gal2022image}, it obtains no prominent gains for more pseudo-tokens and it can serve to efficiently capture new concepts with a high degree of accuracy while using only a single pseudo-word, compared with the same settings of standard textual inversion.

\begin{figure}[t]
\centering
\includegraphics[width=1\columnwidth]{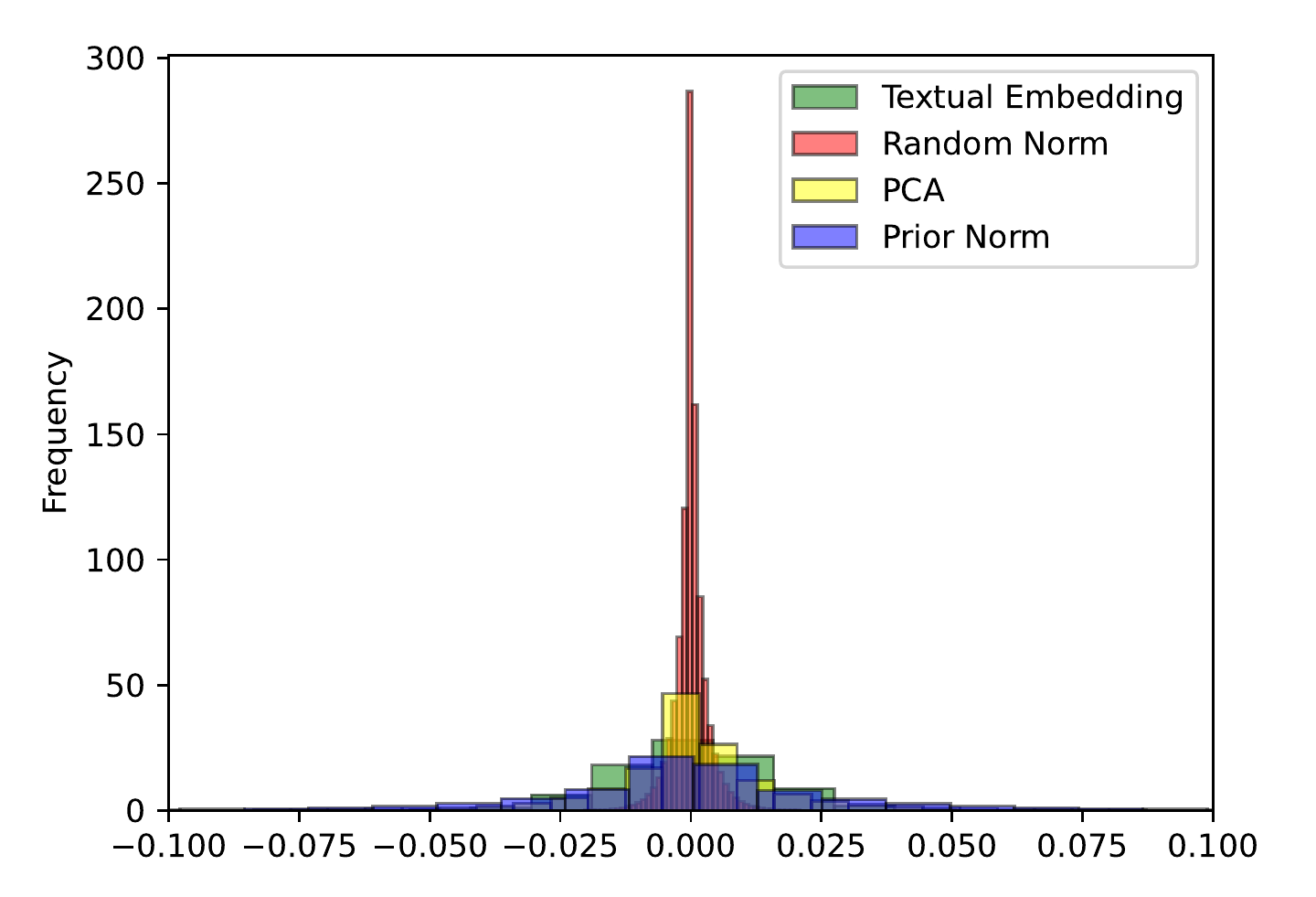} 
\caption{\textbf{Distribution of text embedding in the text-to-image models}. The projection $W_p$ is generated from a decomposition strategy. Our proposed method matches the shape of original text embedding distribution better, leading to faster convergence.
}
\label{fig:distribution}
\end{figure}

\paragraph{Effect of decomposition subspace.}
In this part, we provide a depth analysis of subspace decomposition. 

($\textbf{i}$) The dimensionality of decomposition subspace implicitly balances the trading-off for optimization difficulty and inversion-guided performance. 
We list the subspace dimensionality in the range of $\{64, 256, 512\}$, and we find that: first of all, with the increasing of subspace dimensionality, the editability performance first increases then drops while the reconstruction performance maintains increases slightly. 
Considering both perspectives, we set the decomposition subspace dimensionality to 256 by default. 

($\textbf{ii}$) For decomposition strategies, we find that both PCA and Prior Norm can serve as an effective semantic projection for space disentangling, and achieve a comparable performance with gradient-based inversion. 
To take a closer look into the impact of the projection, we draw the distribution of the initial $W_p Q$. Here, $Q$ is sampled from the normal distribution maintained by the CMA-ES, \emph{i.e.}, set to $\mathcal(0, 0.5)$. By generating $W_p$ from random norm \cite{letham2020re}, PCA, and prior norm, we simulate the distribution of the projected incremental $W_pQ$ and compare it with the distribution of text encoder word embeddings. 
As revealed by Figure \ref{fig:distribution}, when $W_p$ is sampled from the random normal distribution, the projected textual inversion cannot cover the range of word embedding space, and therefore suffers from slow convergence and representation limitations.

\paragraph{Effect of general conditioning initialization. }
We finally compare the GF variants with and without initialization after the same explore steps in an evolution strategy. Figure \ref{fig:ablation} illustrates the superiority of cross-attention based initialization. 
Moreover, we visualize both variants in different evolution steps in Figure \ref{fig:initialization}. 
The cases show that as the optimization step increases, the subject in the generated images becomes more related to the input conceptions and more clear to the prompt.
Meantime, with the incorporation of general conditioned initialization, the model performance at 5000 optimization steps is comparable to result of model with random initialization at 10000 optimization steps, which demonstrate the effectiveness of general condition initialization again.

\subsection{Human Evaluation.}

To better understand how satisfactory the generated images from different methods are, we also conduct a human study to compare our gradient-free inversion against two approaches, \emph{i.e.}, standard textual inversion and directly gradient-free optimization in the original space. 8 evaluators are invited and a subset of 50 concept sets are selected. Each concept is inferred with 20 samples by the model, respectively. 
All the evaluators are provided with three images from a concept set, and ask about the results produced by three models: are the result images corresponding to the concept set? 
From the evaluator's responses, we calculate the metric the percentage of generated images that pass the Turing Test. 
The results of metrics of our method, standard textual inversion, and GFO in original space are 88.2\%, 81.3\%, and 57.6\%, where we believe that the high-dimentional searching space leads to the inferior performance of GFO. 
Overall, our method is clearly the best in terms of human perception criteria.

\section{Conclusion}

This paper proposes a gradient-free textual inversion framework for personalized text-to-image generation, where the gradients can not be accessed and back-propagated to update the parameters. 
In particular, our approach involves an advanced evolution strategy to iteratively optimize the pseudo-token embedding, and such a setting provides more stability and flexibility in practical applications. 
General condition initialization for pseudo-token embedding and decomposed subspace for incremental part is used to accelerate and improve the convergence of iterative process.
Experiments demonstrate that gradient-free optimization is comparable with gradient-based methods, indicating its effectiveness.
We leave the investigation of better exploitation and exploration balancing as future work.

\section*{Societal Impact}

This paper provides an effective tool for generating personal subjects or styles in new contexts without accessing the model parameters. 
While our method enables more safety employment with text-to-image model usage, like other generative model approaches or content manipulation techniques, the generated results exist undesired bias and malicious parties might try to use such images to mislead viewers.
We hope related researchers continue investigating and revalidating these concerns in the future.

\nocite{langley00}
\bibliography{cite}
\bibliographystyle{icml2022}


\end{document}